\pdfoutput=1
\documentclass{SCIS2021}
\usepackage{color}
\definecolor{green}{rgb}{0, 0.5, 0}
\definecolor{orange}{rgb}{0.8, 0.6, 0.2}
\definecolor{red}{rgb}{1.0, 0.0, 0.0}
\definecolor{teal}{rgb}{0.0, 0.4, 0.4}
\definecolor{purple}{rgb}{0.65,0,0.65}
\definecolor{saffron}{rgb}{0.95,0.75,0.2}
\definecolor{turquoise}{rgb}{0.0,0.5,0.5}
\definecolor{turquoise}{rgb}{0.0,0.0,1}

\newcommand{\cy}[1]{{\color{black}{#1}}}
\newcommand{\kx}[1]{{\color{black}#1}}

\newcommand{\zh}[1]{{\color{black}#1}}

\newcommand{\zhn}[1]{{\color{black}#1}}

\newcommand{\nonl}{\renewcommand{\nl}{\let\nl\oldnl}}
\newcommand{\cmark}{\ding{51}}%
\newcommand{\xmark}{\ding{55}}%
\usepackage{multirow}
\usepackage{wrapfig}
\usepackage{mathtools}
\usepackage{amsmath}

\usepackage{amssymb}
\usepackage{amsfonts}
\usepackage{amsopn}
\usepackage{graphicx} 
\usepackage{textcomp}
\usepackage{xfrac}
\usepackage{bbm}
\usepackage{overpic}
\usepackage{subfig}
\usepackage{wrapfig}
\usepackage{multirow}

\usepackage{color}
\definecolor{green}{rgb}{0, 0.5, 0}
\definecolor{orange}{rgb}{0.8, 0.6, 0.2}
\definecolor{red}{rgb}{1.0, 0.0, 0.0}
\definecolor{teal}{rgb}{0.0, 0.4, 0.4}
\definecolor{purple}{rgb}{0.65,0,0.65}
\definecolor{saffron}{rgb}{0.95,0.75,0.2}
\definecolor{turquoise}{rgb}{0.0,0.5,0.5}
\definecolor{brown}{rgb}{0.5, 0.16, 0.16}

\usepackage{overpic}
\usepackage{currfile} 

\newlength\savedwidth
\newcommand\whline[1]{\noalign{\global\savedwidth\arrayrulewidth
		\global\arrayrulewidth #1} %
	\hline
	\noalign{\global\arrayrulewidth\savedwidth}}

\newcommand{\supl}[1]{{\color{black}\emph{#1}}}

\definecolor{lightgray}{rgb}{0.6, 0.6, 0.6}

\usepackage[normalem]{ulem}

\newcommand{\hidecomment}[1]{}

\usepackage{xspace}

\begin{document}
\ArticleType{RESEARCH PAPER}
\Year{2020}
\Month{}
\Vol{}
\No{}
\DOI{}
\ArtNo{}
\ReceiveDate{}
\ReviseDate{}
\AcceptDate{}
\OnlineDate{}

\title{Learning Practically Feasible Policies\\ for Online 3D Bin Packing}{Learning Practically Feasible Policies for Online 3D Bin Packing}

\author[1]{Hang ZHAO$^\dagger$}{}
\author[1]{Chenyang ZHU$^\dagger$}{}
\author[1]{Xin XU}{}
\author[2]{Hui HUANG}{}
\author[1]{Kai XU$^\ast$}{{kevin.kai.xu@gmail.com}}

\AuthorMark{Hang Zhao}

\AuthorCitation{Hang Zhao, Chenyang Zhu, Xin Xu, Hui Huang, Kai Xu}

\contributions{Equal contribution.}

\address[1]{National University of Defense Technology, Changsha {\rm 410073}, China}
\address[2]{Shenzhen University, Shenzhen {\rm 518060}, China}


\abstract{
\zhn{
We tackle the Online 3D Bin Packing Problem, a challenging yet practically useful variant of the classical Bin Packing Problem.
In this problem, the items are delivered to the agent without informing the full sequence information. Agent must directly pack these items into the target bin stably without changing their arrival order, and no further adjustment is permitted. Online 3D-BPP can be naturally formulated as Markov Decision Process (MDP). We adopt deep reinforcement learning, in particular, the on-policy actor-critic framework, to solve this MDP with constrained action space.} To learn a practically feasible packing policy, we propose three critical designs. First, we propose an online analysis of packing stability based on a novel stacking tree. It attains a high analysis accuracy while reducing the computational complexity from $O(N^2)$ to $O(N \log N)$, making it especially suited for RL training. Second, we propose a decoupled packing policy learning for different dimensions of placement which enables high-resolution spatial discretization and hence high packing precision.
Third, we introduce a reward function \zhn{that} dictates the robot to place items in a far-to-near order and therefore simplifies the collision avoidance in movement planning of \zhn{the} robotic arm. Furthermore, we provide a comprehensive discussion on several key implemental issues. \zhn{The extensive} evaluation demonstrates that our learned policy outperforms the state-of-the-art methods significantly and is practically usable for real-world \zhn{applications}.}



\maketitle

\section{Introduction} \label{sec:intro}

\zhn{The bin packing problem (BPP) is one of the most famous problems in combinatorial optimization. It aims to pack a collection of items with various weights into the minimum number of bins.}
The total weight of items in each bin is below the bin's capacity $c$~\cite{korte2012bin}. BPP is a classic NP-hard problem. We are interested in its 3D variant, i.e., 3D-BPP~\cite{martello2000three} \zhn{which introduces much more solving complexity. Given item $i$ with 3D ``weight'' about length, $l_i$, width $w_i$, and height $h_i$, 3D-BPP coordinates planning item's assignment in three dimensions simultaneously. Each 3D dimension has its capacity  including $L \geq l_i$, $W \geq w_i$, and $H \geq h_i$. It is assumed that $l_i, w_i, h_i, L, W, H \in Z^+$. 3D-BPP also pursues to pack the set of items $\mathcal{I}$ with the fewest bins.
}

\begin{figure}\centering
	\includegraphics[width = 0.82\linewidth]{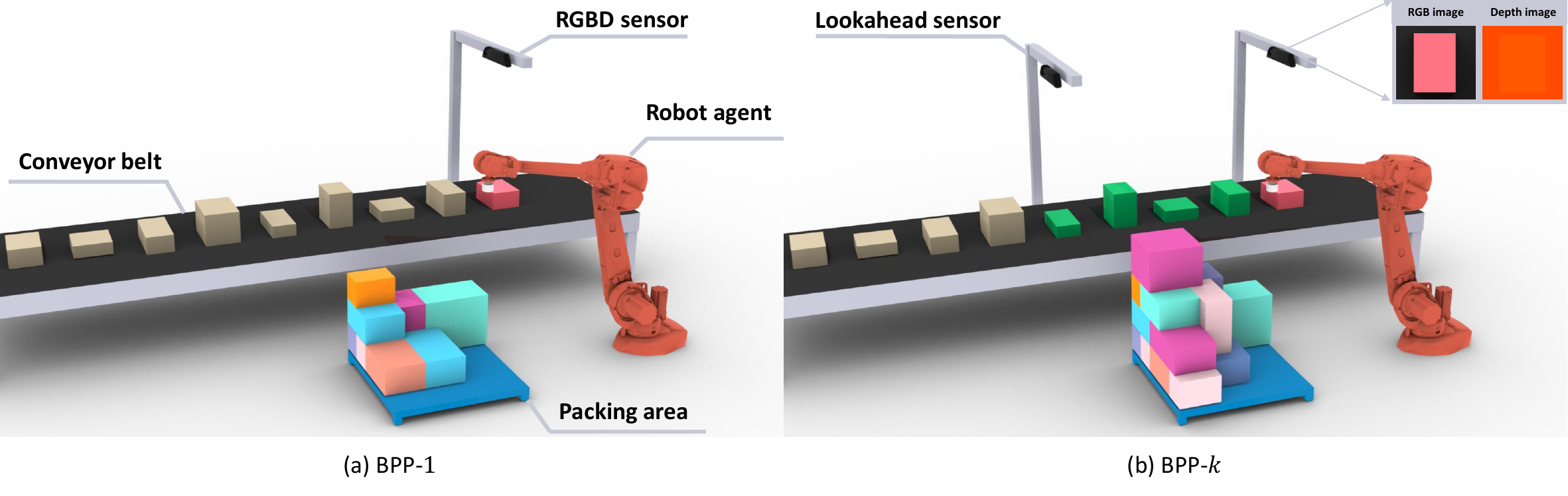}
	\caption{ \zhn{Online 3D-BPP has widely practical applications in logistics, manufacture, warehousing, etc. Left: The agent can only observe the next item to be packed (shaded in red). Right:
	More items (shaded in green) can be observed with additional sensors.
	}	
	}\label{fig:teaser}
\end{figure}

3D-BPP finds widely practical \zhn{applications} in modern packaging, logistics\zhn{,} and manufacturing. It is especially a core technique in developing palletizing robots for intelligent logistics (Figure.~\ref{fig:teaser}).
A palletizing robot is designed to pack boxes into rectangular bins of standard dimension.
\zhn{Maximizing the storage use of bins improves production efficiency like inventorying, wrapping, transportation, and warehousing.} Due to its computational complexity, 3D-BPP is relatively less explored than 1D-BPP. \zhn{Especially when the problem scale increases, the exact algorithm (either using integer linear programming or branch-and-bound) cannot give a solution to the problem within a limited time. Solving the medium-scale 3D-BPP still has to resort to heuristic algorithms~\cite{crainic2008extreme,karabulut2004hybrid}. }

In many application scenarios, an even more difficult problem setting, \emph{Online 3D-BPP}, is highly demanded.
In online 3D-BPP,
\zhn{the information of the full item sequence is not provided to the agent/robot (similar to Tetris).
As shown in Figure~\ref{fig:teaser}, where a robot packs continuously coming parcels online. The RGB-D sensors placed around the robot can only provide a partial vision of the item sequence due to the limited camera view field. Since the conveyor is forwarding sequentially during the packing, only a small period can be given for the robot to unload and place the parcel.}
Such constraints make online 3D-BPP not a pure combinatorial optimization problem since the optimal solution cannot be obtained by brute-force enumeration.

We approach online 3D-BPP by formulating it as a sequential decision making problem and solving it with deep reinforcement learning, similar to~\cite{zhao2020online}. Albeit being quite effective, several limitations of~\cite{zhao2020online}, such as heuristic analysis of packing stability, limited resolution of spatial discretization, and collision agnostic of packing scheme, hinder the practical applicability of learned policy. To this end, we propose the following substantial enhancements to learn practically more feasible policies for online 3D-BPP:
\begin{itemize}
  \item \emph{First}, we impose a physically plausible yet highly efficient stability analysis to facilitate the learning of stable packing policies. In particular, we propose an online analysis of packing stability with a novel stacking tree. Stacking tree analysis attains a $99.9\%$ accuracy of stability analysis with an $\mathcal{O}(N\log N)$ complexity ($N$ is item count). It results in several magnitudes of acceleration of stability analysis over traditional static equilibrium analysis methods which are $O(N^2)$, making it especially suited for RL training. Meanwhile, the accurate analysis also leads to \zhn{a} significant boost of space utilization (by about $10\%$) over~\cite{zhao2020online} based on hand-designed, conservative feasibility evaluation.
  \item \emph{Second}, to deal with large action space caused by high-resolution bin discretization for accurate packing, we propose a novel decoupled packing policy learning scheme. It learns three packing policies for the length and the width dimensions, and the horizontal orientation of the item to be packed, respectively. The three policies are learned with conditionally probabilistic dependency between each other. This enables our method to work with up to $100\times 100$ discretization resolution which is intractable for \zhn{the} method in~\cite{zhao2020online}.
  \item \emph{Third}, to learn a more practically usable policy, we introduce a reward function which encourages the robot to place items in a far-to-near order. This greatly eases the collision avoidance between the robot and the packed items and simplifies the movement planning of the robotic arm.
\end{itemize}

Our method is formulated as a constrained Markov decision process (CMDP)~\cite{altman1999constrained} and \zhn{adopts} the on-policy actor-critic framework~\cite{mnih2016asynchronous,wu2017scalable}. Different from~\cite{zhao2020online}, we compute the feasibility mask for the placement actions based on the stacking tree analysis. The feasibility mask is then \zhn{provided} to the actor networks and then used to modulate the action probabilities output by the actor.
We also discuss several practical issues in the real robot implementation of our algorithm.
Finally, we conduct extensive \zhn{evaluations} to validate the efficacy of the new designs.


\section{Related Work}\label{sec:related}

\zhn{\paragraph{Bin Packing Problem (BPP)} is a long-term 
concern in combinatorial optimization, and the earliest literature can be traced back to the sixties~\cite{kantorovich1960mathematical}. The most typical bin packing problem is 1D-BPP which seeks for an assignment of a collection of items with various scalar weights to multiple bins and minimizes the used bin number. Knowing to be strongly NP-hard, most existing literature focuses on designing good heuristic and approximation algorithms and their worst-case performance analysis~\cite{coffman1984approximation}. Bin packing problem exists many variants, 2D- and 3D-BPP are natural generalizations of them. For high-dimension BPP, an item has a high-dimension size of \zhn{the} width, height, and/or depth, which  differentiates the verification of the packing feasibility. The complexity and the difficulty significantly increase for high-dimension BPP instances. 
According to the timing statistic reported in~\cite{martello2000three}, exactly solving 3D-BPP of a size matching an actual parcel packing pipeline remains infeasible. 
There have been several strategies in designing fast approximate algorithms, e.g., guided local search~\cite{faroe2003guided}, greedy search~\cite{de2003greedy}, and tabu search~\cite{lodi1999approximation,crainic2009ts2pack}. In contrast, genetic algorithms \zhn{lead} to better solutions as a global, randomized search~\cite{li2014genetic,takahara2005evolutionary}.

}

Similar \zhn{strategies have} also been adapted to Online BPP works like ~\cite{ha2017online,wang2016benchmarking,zhao2020online, hong2020smart}. Different from the offline setting, the size information of coming items is unknown for the agent to optimize the packing and the packing order can not be adjusted as well. It makes Online BPP a much more challenging problem. Some works ~\cite{ha2017online,wang2016benchmarking} have employed the hand-coded heuristics based on the
human experience. Meanwhile, works like ~\cite{zhao2020online} have adopted deep reinforcement learning to learn how to pack things effectively through trials and error optimization.

\paragraph{Stability Estimation} is essential in \zhn{the} bin packing problem. However, most previous work focuses on \zhn{the 3D BPP} \zhn{ignores} it in their solution due \zhn{to} the high computational load. Estimating \zhn{the} stability of stack objects for physics \zhn{engines} is widely studied in past decades. \cite{erleben2007velocity} introduces novel complementarity formulation, \zhn{solver,} and error correction algorithms for collision detection framework which can enable the stability estimation running in real-time for large scale objects. In contrast, \cite{thomsen2015simulating} presents a real-time physics engine to improve structured
stacking behavior with small-scale objects. \cite{hsu2012automated} present a \zhn{constraint-based} method to stabilize a stack of piles. \cite{han2013believability} introduces a hypothesis for \zhn{physical} simulation simplification which is freezing transformations of objects in a random pile does not affect the visual plausibility of a simulation. However, none of them can be efficient enough for stability estimation in a DRL training framework.


\paragraph{Deep Reinforcement Learning (DRL)}
\zhn{
has demonstrated tremendous success in learning complex behavior skills and solving challenging control tasks with high-dimensional raw sensory state-space~\cite{lillicrap2015continuous,mnih2015,mnih2016asynchronous}.
The success can be attributed to the utilization of high-capacity deep neural networks for powerful feature representation learning and function approximation.
Since the seminal work of deep Q-network (DQN)~\cite{mnih2015}, a large body of literature has emerged. The existing research is largely divided into two lines: value function learning~\cite{mnih2015,wang2015dueling} and policy search~\cite{silver2014deterministic,barth2018distributed}.
Actor-critic methods, designed to combine the two approaches, have grown in popularity.
Asynchronous Advantage Actor-Critic (A3C)~\cite{mnih2016asynchronous} is a representative actor-critic method.
It combines advantage updates with the actor-critic formulation and adopts asynchronously updated
policy and value function networks trained in parallel with multiple agents.
In A2C~\cite{schulman2017proximal}, on the other hand, the networks of multiple agents are updated synchronously.

We base our actor-critic architecture on ACKTR~\cite{wu2017scalable} which applies trust-region optimization to both the actor and the critic.
We propose a series of adaptions for solving online 3D BPP.
To realize constrained reinforcement learning, we propose a simple approach by projecting the trajectories sampled from the actor to the constrained state-action space. To enable our agent to fit the high-resolution demand for real packing, we decompose the actor into three actor-heads and get them to predict the corresponding actions in sequence. We also feed the last predicted action into the next actor-head as a conditional probabilistic dependency to achieve a more stable training process.
}

\zhn{\paragraph{RL for Combinatorial Optimization}
has booming development in recent years. 
Bello et al.~\cite{bello2016neural} combine \emph{RL pretraining} and \emph{active search} and demonstrate that RL-based optimization outperforms \zhn{the} supervised learning framework when tackling NP-hard combinatorial problems.
Kool et al.~\cite{attention2019route} encode TSP graphs with 
Transformer ~\cite{attention2017need} and train this model with RL method.  
Zhang et al.~\cite{ZhangSC0TX20} enable an end-to-end RL agent to master priority dispatching rules for solving Job-shop scheduling problem. Wang et al.~\cite{WangLY20} apply RL to automatically generating a move plan for indoor scene arrangement. \cite{hu2017solving} and  ~\cite{laterre2018ranked}
are both devoted to solving offline 3D-BPP where the main goal is to find an optimal sequence of items.

}

\section{Method}\label{sec:method}

\begin{figure*}
  \centering
  \includegraphics[width=\linewidth]{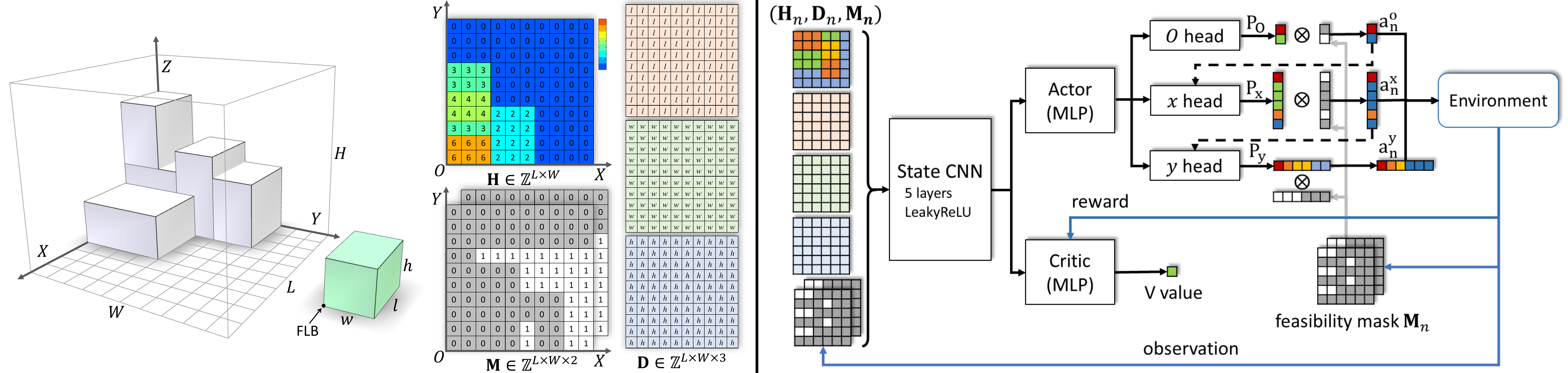}
  \caption{
The environment state of the agent. The grey boxes indicate the items already packed, which also represents the bin configuration. The green box is the next item to be packed and it can only be placed at the grid cell where feasibility mask $\mathbf{M}$ is 1.
Right: The network architecture \zhn{with decomposed actor-heads}. Note that the training of three actor-heads \zhn{is} coupled with conditional probabilistic dependencies, \zhn{actor-heads perform their prediction tasks in sequence}.}
 \label{fig:3D-BPP}
\end{figure*}

We adopt a similar problem configuration for online 3D-BPP as \cite{zhao2020online}. Each item $n\in\mathcal{I}$ is a cube whose size is $[l_i, w_i, h_i]$. As soon as an item arrives the packing area, the agent needs to place it to the bin immediately while only being aware of the information of next few coming items $\mathcal{I}_o\subset\mathcal{I}$. Our method is implemented based on the architecture of constrained DRL and we will start our problem statement and formulation under the context of DRL.


\subsection{Problem Statement and Formulation}
The DRL formulation of our method can be expressed as $(\mathcal{S}, \mathcal{A}, P, R)$, which is a Markov decision progress. $\mathcal{S}$ is constructed with a set of environment states $\mathcal{S}$, the action set $\mathcal{A}$, reward function $R : \mathcal{S} \times \mathcal{A} \rightarrow \mathbb{R}$, and transition probability function $P : \mathcal{S} \times \mathcal{A} \times \mathcal{S} \rightarrow [0,1]$.
$P(s'|s,a)$ gives the probability of transiting from $s$ to $s'$ for given action $a$.
Our method is model-free since we do not learn $P(s'|s,a)$ explicitly. The policy $\pi : \mathcal{S} \rightarrow \mathcal{A}$ is a map from states to probability distributions over actions, with $\pi(a|s)$ denoting the probability of selecting action $a$ under state $s$. For DRL, we seek for a policy $\pi$ to maximize the accumulated discounted reward, $J(\pi)=E_{\tau \sim \pi}[\sum_{t=0}^{\infty}\gamma^tR(s_t,a_t)]$. Here, $\gamma \in [0,1]$ is the discount factor, and $\tau = (s_0,a_0,s_1,\ldots)$ is a trajectory sampled based on the policy $\pi$.

\zhn{
\paragraph{Environment State}
A complete 3D-BPP state representation should include the following three parts: the current configuration of the bin, the coming items to be placed, and the feasibility mask. To parameterize the bin configuration, we discretize its bottom area as an $L \times W$ regular grid along the length ($X$) and the width ($Y$) directions, respectively. We record at each grid cell the current height of stacked items, leading to a 2D integer \emph{height map} $\mathbf{H}_n\in\mathbb{Z}^{L \times W}$ (see Figure~\ref{fig:3D-BPP}). The dimensionality of item $n$ is given as  $\mathbf{d}_n = [l_n, w_n, h_n]^\top\in\mathbb{Z}^3$.
The feasibility mask $\mathbf{M}_{n,o}$ is a binary matrix of size $L \times W$ indicating the placement feasibility of $n$ with orientation $o$ at each grid cell. Putting together, the current environment state can be written as $s_n = \{\mathbf{H}_n, \mathbf{d}_n, \mathbf{d}_{n+1},...,\mathbf{d}_{n+k-1}, \mathbf{M}_{n,o} \}$.
We first consider the case where $k=|\mathcal{I}_o|=1$ (Figure~\ref{fig:teaser} (a)), and name this special instance as \textbf{BPP-$1$}.
In other words, \textbf{BPP-$1$} only considers the immediately coming item $n$ i.e., $\mathcal{I}_o=\{ n \}$. We then generalize it to \textbf{BPP-$k$} with $k>1$ (Figure~\ref{fig:teaser} (b)) afterwards.
}

\paragraph{Action and State Update} We consider only horizontal, axis-align orientations of an item, which means that each item $n$ has two possible orientations $o_n(d_n)=\{[l_n, w_n, h_n]^\top, [w_n, l_n, h_n]^\top\}$. During the packing, the agent places orientated $n$'s front-left-bottom (FLB) corner (Figure~\ref{fig:3D-BPP} (left)) at a certain grid cell or the loading position (LP) in the bin. For instance, if the agent chooses to put $n$ at the LP of $(x_n, y_n)$ with the orientation adjustment $o_n$, this action is represented as $a_n = (x_n, y_n, o_n)$.
The range of $\mathcal{A}$ would increase dramatically with \zhn{a} large resolution of $x_n, y_n$. In other words, it would be difficult to optimize action while high place accuracy is needed. To solve this problem, we propose an action space decomposition method in Section~\ref{sec:network_architecture}.
After $a_n$ is executed, $\mathbf{H}_n$ is updated by adding $h_n$ to the maximum height over all the cells covered by $n$:
$\mathbf{H}'_{n}(x, y, o) = h_{\text{max}}(x, y, o) + h_n$ for $x\in[x_n, x_n + (1-o_n)l_n+o_nw_n], y\in[y_n, y_n + (1-o_n)w_n+o_nl_n]$,
with $h_{\text{max}}(x, y)$ being the maximum height among those cells.

\zhn{
\paragraph{Feasibility Constraint}
A practical online BPP solution should also premeditate the stability of a placement besides securing enough valid space for future items.
Placing an item in a risky LP will end the packing episode early and even cause economic damage in practice.
We propose a stacking tree based stability estimation method in Section~\ref{sec:stability} to form feasibility masks (see Figure~\ref{fig:3D-BPP}) as optimization constraints to avoid insecure behaviors.  }
\zh{Our problem becomes a constrained Markov decision \zhn{process} (CMDP)~\cite{altman1999constrained} in this scenario. Typically, one augments the MDP with an auxiliary cost function $C : \mathcal{S} \times \mathcal{A} \rightarrow \mathbb{R}$ mapping state-action tuples to costs, and require that the expectation of the accumulated cost should be bounded by $c_m$:  $J_C(\pi) = E_{\tau \sim \pi}[\sum_{t=0}^{\infty}\gamma_C^tC(s_t,a_t)] \leq c_m$. Several methods have been proposed to solve CMDP based on e.g.,
algorithmic heuristics~\cite{uchibe2007constrained}, primal-dual methods~\cite{chow2017risk}, or constrained policy optimization~\cite{achiam2017constrained}.
While these methods are proven effective, it is unclear how they could fit for 3D-BPP instances, where the constraint is rendered as a discrete mask. In this work, we propose to exploit the mask $\mathbf{M}$ to guide the DRL training to enforce the feasibility constraint without introducing excessive training complexity.}

\subsection{Stability Estimation}\label{sec:stability}
Stack stability estimation is critical in \zhn{the} bin packing problem, especially in 3D. A good stability estimation for LPs of $n$ would not only secure the safety of the placement but also decrease the searching range of the action space. To calculate $\mathbf{M}_{n,o}$, the most straightforward solution is to simulate the force analysis among packed items $\mathcal{I}_{\text{packed}}\subset\mathcal{I}$ while $n$ is placed with orientation $o$. However, this simulation would become extremely complicated while the $|\mathcal{I}_{\text{packed}}|$ increasing since force analysis with dense structured stacking is \zhn{an} NP-hard problem\cite{erleben2007velocity}. To ensure the stack stability estimation for LPs is real-time, we propose a \zhn{centroid-based} approach which \zhn{improves} the efficiency \zhn{by} more than 100 times.

\begin{figure*}[h]
  \centering
  \includegraphics[width=0.8\linewidth]{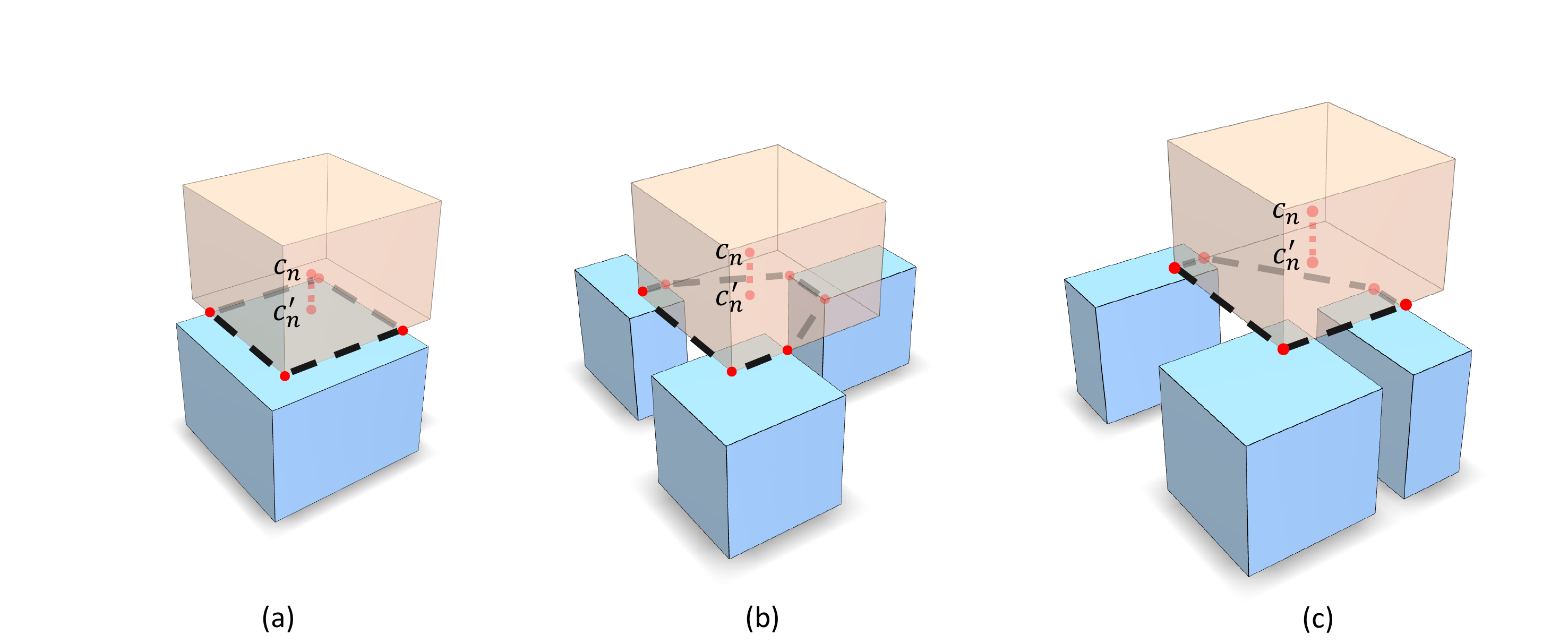}
  \caption{(a): \zhn{The centroid of n is supported by a packed item directly therefore stable. } (b): The vertical projection $c'_n$ of centroid $c_n$ is inside \zhn{the} convex hull which is constructed by contact points of $\mathcal{I}_{\text{support}}$. (c): $c'_n$ falls outside the supported convex hull, $n$ is unstable in this situation.}
 \label{fig:convex}
\end{figure*}

\paragraph{Supported Centroid} In industrial bin packing applications, item $n$ to be packed usually has uniform mass distribution. An item $n$ is stable if its centroid is supported in this scenario. Specifically, \zhn{an} LP of $n$ is considered stable if satisfies any of \zhn{the} following conditions: 1) The centroid $c_n$ of $n$ is directly supported by a packed item with this LP. 2) $n$ is supported by a group of items $\mathcal{I}_{\text{support}}\subset\mathcal{I}_{\text{packed}}$ and $c_n$ is inside the convex hull which is constructed by contact points of $\mathcal{I}_{\text{support}}$, as illustrated in \cy{Figure~\ref{fig:convex}}. The \zhn{centroid-based} discrimination is easy to implement with \zhn{the} boolean operation and can be highly parallelized.

\paragraph{Adaptive Stacking Tree} However, we not only need to estimate the stability of the current LP but also calculate the stability changes of $\mathcal{I}_{\text{packed}}$ which \zhn{is} introduced by the new contact. The mass of $n$ changes the centroids of items \zhn{which} support it, and these changes may alter their stability. In other words, a top-down traverse of stability update is necessary. However, the amount of work done in this fashion is $\mathcal{O}(N^2)$ since a whole traverse is needed for each placed item. We propose an adaptive stacking tree structure to update the mass distribution flow of $\mathcal{I}_{\text{packed}}$ efficiently in $\mathcal{O}(N\log N)$. The key idea here is that the mass of $n$ at LP would only distribute to the items $\mathcal{I}_{\text{active}}\subset\mathcal{I}_{\text{packed}}$ which support it, directly or indirectly. The stability of $\mathcal{I}_{\text{packed}}-\mathcal{I}_{\text{active}}$ would not change, and the stability update should only be done with $\mathcal{I}_{\text{active}}$ but not the whole packed bin.

\begin{figure*}
    \centering
    \includegraphics[width=0.9\linewidth]{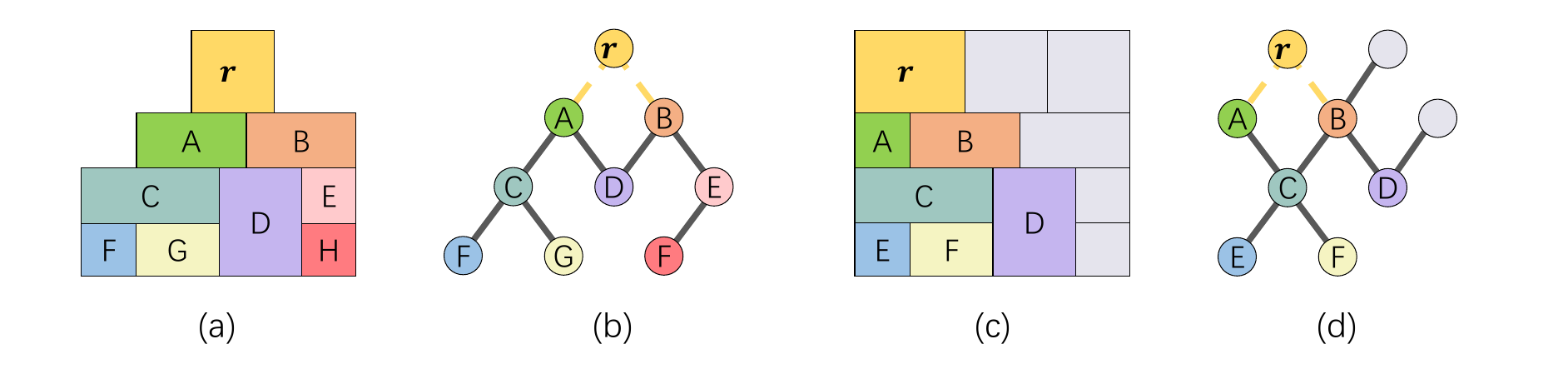}
    \caption{(a): $r$ is the current item to be packed which will trigger an adaptive stacking tree to be updated. (b): The adaptive stacking tree of (a), the mass of $r$ flows along each graph edge. (c): Placement of $r$ does not change the mass flow over items with gray color. (d) The adaptive stacking tree of (c), the mass distribution of colored nodes \zhn{needs} to be updated. Note that the mass of inactive items (gray ones) will distribute to their adjacent active items while the update.}
   \label{fig:stack}
  \end{figure*}

In \cy{Figure~\ref{fig:stack}}, the mass distribution flow of $\mathcal{I}_{\text{packed}}$ can be formed as a graph $\mathcal{G}$, the nodes represent the items in the bin and the edges indicate the amount of mass distribution between two adjacent items. As we discussed above, only a subgraph $\mathcal{G}_n$ of $\mathcal{G}$ \zhn{needs} to be updated while $n$ is placed at LP, where $\mathcal{G}_n$ only contains items in $\mathcal{I}_{\text{active}}$. Obviously, $\mathcal{G}_n$ is a tree which we denote as the adaptive stacking tree. $\mathcal{G}_n$ can be constructed from $n$ with a top-down fashion, and only edges of $\mathcal{G}_n$ \zhn{need} to be updated while $n$ is placed. The mass distribution update is processed from the root $n$ to the bottom in $\mathcal{G}_n$ based on the calculated mass distribution, while the edges in $\mathcal{G}-\mathcal{G}_n$ should stay unchanging. This process can be easily implemented during packing in an incremental fashion based on a linked list data structure, which is both memory and time efficient.

\paragraph{Stability Update} Updating mass distribution in $\mathcal{G}_n$ is critical for stability estimation. However, it would be highly time-consuming if we operate a force simulation here. We formulate a simple but effective approach based on the principle of leverage, which can achieve $99.9\%$ accuracy with 100 times efficiency. While $n$ is placed at LP, three types of mass distribution would be discussed next. If $n$ is only supported by a single item $n_0$, the whole mass $m_n$ of $n$ and items above it would transport to $n_0$. The new centroid of the group of $\{n,n_0\}$ would change to,

\begin{equation} \label{eq:centroid}
	c_{\{n,n_0\}}=\frac{c_n\cdot m_n+c_{n_0}\cdot m_{n_0}}{m_n+m_{n_0}}
\end{equation}

\begin{wrapfigure}{r}{0.38\linewidth}\centering
\vspace{-20 pt}
\hspace{0 pt}\includegraphics[width=\linewidth]{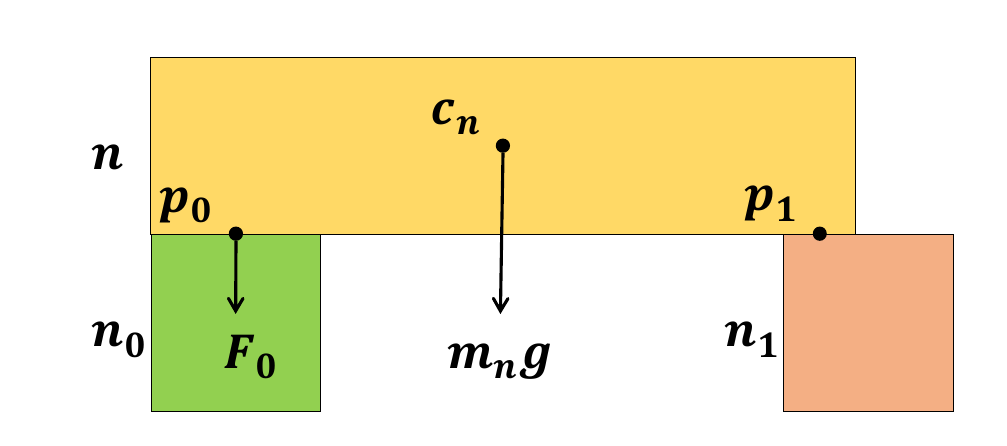}
\end{wrapfigure}

Note that we only calculate centroid in \zhn{the} $XY$ plane since the value along \zhn{the} $Z$ axis would not influence the stability. If $n$ is supported by two \zhn{items} $n_0$ and $n_1$ safely, we assume the mass distribution of $m_n$ would obey the principle of leverage. As shown in right, $n_0$ needs to undertake $F_0$ to make the whole system stable.

\begin{equation} \label{eq:distribution}
	F_0 = g\cdot \|(c_n-p_1)\times m_n\|/\|p_0-p_1\|
\end{equation}

\begin{equation} \label{eq:newmass}
M\{n,n_0\}=F_0/g+m_{n_0}
\end{equation}

where $p_0$ and $p_1$ are two contact points of $n_0$ and $n_1$. We can also calculate $F_1$ similarly. Then the centroid of group \{$n,n_0$\} and \{$n,n_1$\} can be updated with Equation(\ref{eq:centroid}) as well as the mass flow $M\{n,n_0\}$ and $M\{n,n_1\}$ which distributed to $n_0$ and $n_1$. If $n$ is safely supported by more than two items $\{n_k\}$, the mass distribution of $m_n$ would be calculated with Equation(\ref{eq:distribution}) for each two items in $\{n_k\}$. Then we will have $C_k^2+1$ constraints for mass distribution, a least squares optimization would be adopted to find the optimal $F_i$ for each item in $\{n_k\}$. The centroid of group $\{n,n_i\}$ can be updated as well. Then each group $\{n,n_i\}$ would trigger the next iteration of update as well as $n$, this process would stop if the update \zhn{reaches} the bin bottom or \zhn{instability} is detected.

\zh{
\subsection{Network Architecture \zhn{and Training Method}}\label{sec:network_architecture}
}

\zhn{
We adopt ACKTR~\cite{wu2017scalable}, which is a state-of-the-art on-policy framework.
It iteratively updates an actor and a critic using Kronecker-factored approximate curvature (K-FAC) \cite{martens2015optimizing} with trust region. The actor is trained to learn a policy network that outputs the probability of choosing each action (i.e., placing $n$ at each LP).} The critic trains a state-value network predicting the state value $\displaystyle V(s_{n})$ to indicate how much reward is earned from $\displaystyle s_n$, it is adopted to train our actor network. \zhn{\cite{zhao2020online} has demonstrated that ACKTR has a surprising superiority over other DRL algorithms like SAC~\cite{haarnoja2018soft}.}

\zhn{The BPP state consists of three components, the height map $\mathbf{H}_n$, the coming items to be placed $\mathbf{D}_n$, and the feasibility mask $\mathbf{M}_{n,o}$.  We use a Convolutional Neural Network (CNN) to encode the raw BPP state.
For calculation convenience, we ``stretch'' $\mathbf{d}_n$ into a three-channel tensor $\mathbf{D}_n \in \mathbb{Z}^{L \times W \times 3}$ so that each channel of $\mathbf{d}_n$ spans an $L \times W$ matrix with all of its elements being $l_n$, $w_n$, or $h_n$, respectively (also see Figure~\ref{fig:3D-BPP} (left)). Consequently, state $s_n = (\mathbf{H}_n, \mathbf{D}_n, \mathbf{M}_{n,o})$ becomes an $L \times W \times 6$ array (Figure~\ref{fig:3D-BPP} (right)).
}

\paragraph{Reward Shaping}

\zhn{We want the agent to learn practical skills which meet the demands of both safety and efficiency without influenced
by various robot type. The efficiency need is satisfied by the reward term which is about the volumetric occupancy introduced by the current item: $10 \times l_n \cdot w_n \cdot h_n / (L \cdot W \cdot H)$ for item $n$. \cite{zhao2020online} reports that the step-wise reward is superior to a termination one (e.g. the final space utilization).}

We found that if the agent is only trained with $r_n$, the actor network would intend to place items uniformly on the $XY$ plane of the bin. However, the robot arm is usually equipped aside rather than above the bin in practice, packing \zhn{items} near the robot arm first would potentially introduce collisions as \zhn{shown} in {Figure~\ref{fig:collision}. To avoid it, \cite{martello2007algorithm} propose that the packing should start from the farther corner as possible. Nonetheless, the performance would drop significantly if we force the actor \zhn{to follow} this principle strictly. To balance the two factors, we introduce a side reward to give our \zhn{agent} a soft constraint.

\begin{figure*}
    \centering
    \includegraphics[width=0.8\linewidth]{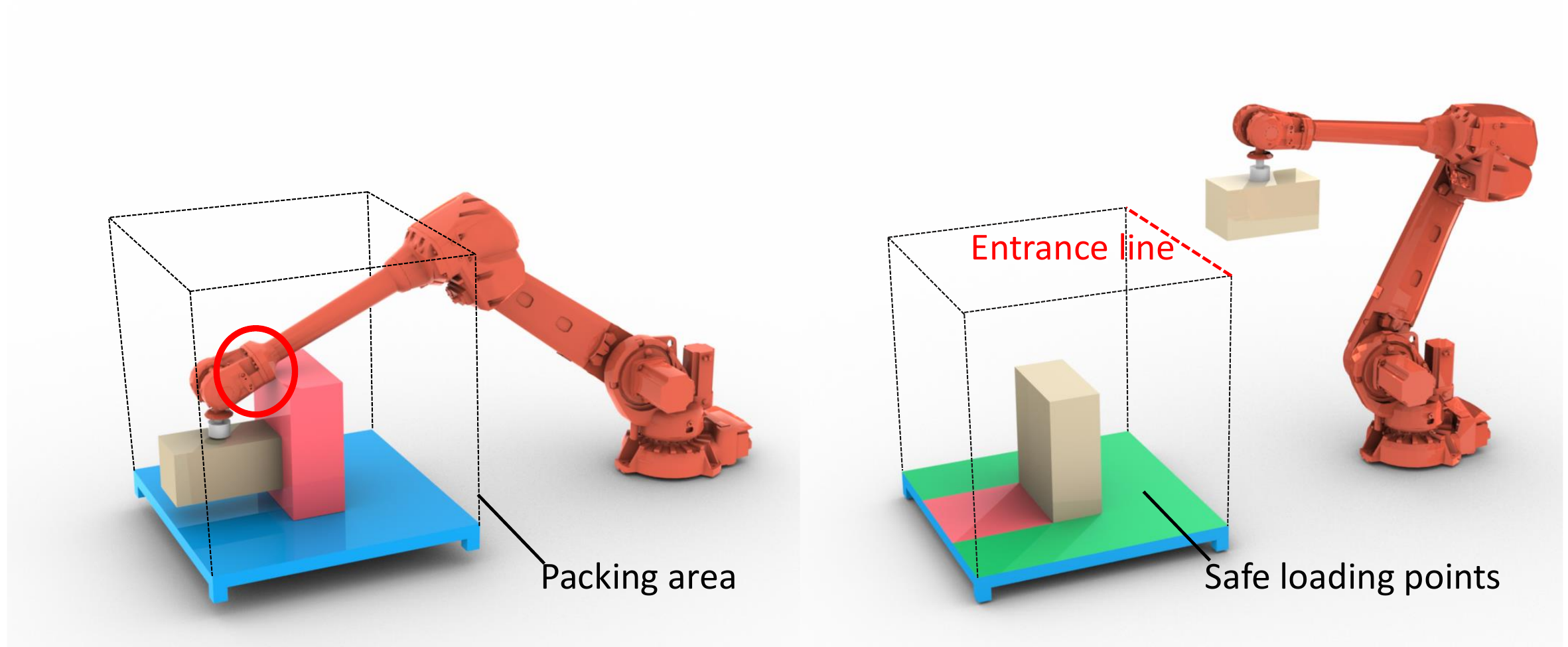}
    \caption{Left: The red box packed in the middle of the bin first \zhn{introduces} potential collisions for \zhn{the} following packing. Right: If the robot always \zhn{enters} the packing area at the same \zhn{entrance} line, packing on the green area may introduce fewer potential collisions.}
   \label{fig:collision}
  \end{figure*}

We assume that the robot arm would always enter the packing area at the same \zhn{entrance} line \zhn{(EL)} \cy{(Figure~\ref{fig:collision})}. If there is no obstacle on the straight line from EL to a feasible LP, \zhn{the} packing item at this LP can hardly introduce collisions and this type of LP is a safe LP. In other words, the more LPs which satisfy this principle the more likely our next packing is collision-free. We denote the sum of upon volume of all safe LPs as $V_{\text{safe}}$\cy{(Figure~\ref{fig:collision})}, our side reward is formulated as $r^\prime_n = V_{\text{safe}}/(L \cdot W \cdot H)$ to ensure the maximum amount of safe LPs. The final reward of {the} system is then revised as $r_n = \alpha  \times l_n \cdot w_n \cdot h_n / (L \cdot W \cdot H) + \beta V_{\text{safe}}/(L \cdot W \cdot H)$, where $\alpha = 10 \text{ and } \beta = 0.1$ since the packing performance is still the primary. \zhn{The packing episode ends when the current item is not placeable and the $r_n$ is zero.}

\paragraph{Action Space Decomposition} The resolution choice of $\mathbf{D}_n$ would significantly influence the range of action space $\mathcal{A}$. However, we need \zhn{a} high resolution of $\mathbf{D}_n$ to ensure the packing accuracy in practical applications which would dramatically enlarge $\mathcal{A}$. To solve this problem, we propose a multi-task training method. The action prediction task is decomposed into three subtasks, predict $x_n$, $y_n$ and $o_n$ respectively \cy{(Figure~\ref{fig:3D-BPP})}. The decomposition would reduce the action space from $O(n^3)$ to $O(3n)$. Different from previous work~\cite{tavakoli2018action} which learns state-action value functions separately for each task, we train different actors for each task and use unified critic function $\displaystyle V(s_{n})$ for the three individual actors to predict $P_x, P_y, P_o$. Note that the training of three actors \zhn{is} coupled as well, the advantage is that the $x_n$, $y_n$\zhn{,} and $o_n$ can be predicted in a correlated fashion which is reasonable in our problem configuration. The policy gradient for actor network parameters training to predict $P_x, P_y, P_o$ is formulated as:

\begin{equation} \label{eq:gradient}
	\nabla \theta _{\text{actor}}=(r_n+\gamma V(s_{n+1})-V(s_n))\nabla\log P_{\text{actor}}(a_n|s_n)
\end{equation}

where $\theta _{\text{actor}}, \text{actor}\in\{x,y,o\}$ is the tunable parameters for different actors, $\gamma \in [0,1]$ is the discount factor and we set $\gamma$ as 1 so that $\displaystyle V(s_n)$ can directly present how much reward can agent obtain from $s_n$ on.

\paragraph{Loss function} The whole network is trained via a composite loss. The actor loss $L_{actor}$ is inferred from Equation(\ref{eq:gradient}) and the critic loss is constructed based on our reward function $r_n$, which are the loss functions used for training the actor and the critic, respectively. Next, we use the feasibility mask $\mathbf{M}_n$ given by stability estimation to modulate outputs of three actors in order, i.e., the probability
distribution of the actions. In theory, if the LP at $(x, y)$ is infeasible for $n$ with orientation adjustment $o$, the corresponding probability $P(a^o_n= o|s_n), P(a^x_n= x|s_n), P(a^y_n= y|s_n)$ should be set to $0$ respectively. However, we find that setting $P$ to a small positive quantity like $\epsilon = 10^{-20}$ works better in practice -- it provides a strong penalty to an invalid action but a smoother transformation beneficial to the network training.
Our loss function is defined as:

\begin{equation} \label{eq:loss_function}
  L = \alpha \cdot L_{\text{actor}} + \beta \cdot L_{\text{critic}} + \omega \cdot E_{\text{inf}} - \psi \cdot E_{\text{entropy}}.
  \end{equation}

  \begin{equation}
    \left\{
    \begin{array}{ll}
      \displaystyle L_{\text{actor}} & \displaystyle =  (r_{n}+ \gamma  V(s_{n+1})-V(s_n))\log
    P_{\text{actor}}(a_n | s_n)\\
    \displaystyle {E_{\text{inf}}} &\displaystyle =\sum_{{\mathbf{M}_{n,o}(x,y)=0}} P_{\text{actor}}(a_n|s_n)\\
    \displaystyle E_{\text{entropy}} & \displaystyle= \sum_{\mathbf{M}_{n,o}(x,y)=1}-P_{\text{actor}}(a_n |s_n) \cdot
    \log \big(P_{\text{actor}}(a_n |s_n)\big)\\
    \displaystyle L_{\text{critic}} & \displaystyle = (r_{n}+ \gamma  V(s_{n+1})-V(s_n))^2
    \end{array}
    \right.
    \end{equation}	

where $\text{actor}\in\{x,y,o\}$. To further discourage infeasible actions, we explicitly minimize the summed probability at all infeasible LPs with $E_{\text{inf}}$. Meanwhile, the action entropy loss $E_{\text{entropy}}$ is adopted to push the agent to explore more LPs. In this way, we stipulate the agent to explore only feasible actions.
We recommend the following parameters which lead to consistently good performance throughout our tests:
$\alpha=1$, $\beta=0.5$, and $\omega=\psi=0.01$.

\zh{
\subsection{BPP-$k$ with $k=|\mathcal{I}_o|>1$}
\label{sec:reorder}
}

\zhn{In a more general case, the agent receives the information of $k>1$ \emph{lookahead} items (i.e., from $n$ to $n+k-1$). With additional information embedded into the environment state, the agent is expected to learn the policy $\pi(a_n|\mathbf{H}_n,\mathbf{d}_n,...,\mathbf{d}_{n+k-1})$ and have better performance. \cite{zhao2020online} claims that simply encoding the lookahead information into the network input cannot help. We adopt the search-based solution to enable the agent to leverage lookahead information explicitly.}

\zhn{
\paragraph{Virtual Placement Order}
\cite{zhao2020online} claims that the current item $n$'s placement should be conditioned on the next $k-1$ one.
They enumerate different permutations of the sequence $(\mathbf{d}_n,...,\mathbf{d}_{n+k-1})$ and drive the actor network to give related plans. To evaluate each sequence, they sum up the accumulated reward and the critic value of the end state after the $k$-th item is placed. The most promising $a_n$ can be found in the sequence with the highest evaluation score. Note that only $n$'s placement is determined in one permutation search and the actual placement of the $k$ items still follows the order of arrival. To make the search scalable when $k$ is large, \cite{zhao2020online} adapts the Monte Carlo tree search (MCTS)~\cite{silver2017mastering} to this problem and scales down the computational complexity from $O(k!)$ to $O(km)$ where $m$ is the number of permutations sampled.
}

\zhn{
\paragraph{Adaptions for MCTS}

We have made some adjustments to \cite{zhao2020online}'s MCTS method so that it can be employed in practical scenarios. 
Firstly, we multiply each virtual item's mass with an extremely small coefficient ($10^{-6}$) during the search so that the stability estimation for $n$ will not be influenced by nonexistent items. The mass coefficient cannot be zero otherwise the virtual items will lose physical constraint with other virtual ones. We also noticed that calculating the feasibility mask based on height map update and stability estimation in MCTS is inevitably time-consuming for real-time since this process would be invoked thousands of times in a whole MCTS.
To improve the efficiency, we adpot a root-based parallelization for the MCTS. Different from leaf-based parallelization and tree-based parallelization\cite{chaslot2008parallel}, there is no communication between clones during the parallelization. All the root children of the separate Monte Carlo trees are merged with their corresponding clones after every parallelization iteration. After the search, we choose the action $a_n$ corresponding to the permutation with the highest path value.
MCTS allows a scalable lookahead for BPP-$k$ with a complexity of $O(km)$ where $m$ is the number of permutations sampled.

However, parallelizing MCTS will sacrifice the sampling frequency per thread and harm the performance. 
We modify the MCTS node expansion strategy to ensure that MCTS focuses more on meaningful sequences.
The items coming soon in actual order will be more likely to be sampled during the permutation search since a partial sequence given by a virtual order is sometimes meaningless. The sample possibility of $item(v_i)$ is $P_{\text{sample}}(x)\sim \mathcal{N}(0,1)$ in our implementation where $x$ means $item(v_i)$'s sorted arrival index among unselected items. This attention-based fashion would give MCTS a soft constraint to sample more items with the actual order.
}
\subsection{Implementation }\label{sec:implementation}
We implement our system in a practical industry environment to validate the actual packing performance. Figure~\ref{fig:realrobot} illustrates the online autonomous bin packing system with an RGB-D sensor and a robot arm.

\begin{figure*}[h]
    \centering
    \includegraphics[width=0.9\linewidth]{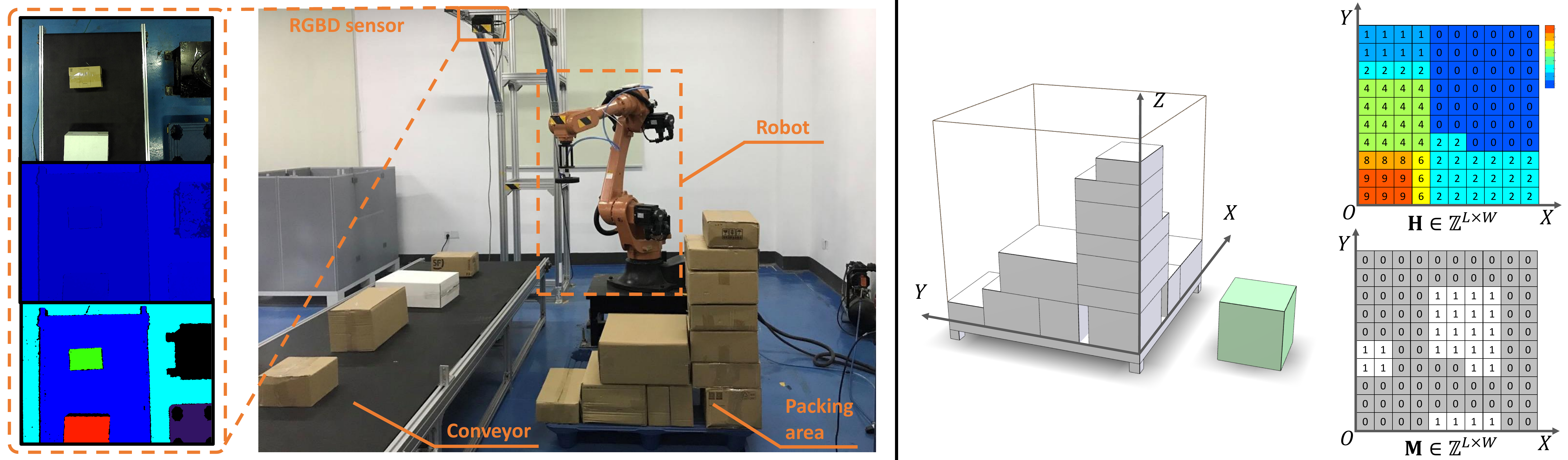}
    \caption{We implement our system in a practical industry environment. Left: The online autonomous bin packing system with an RGB-D sensor and a robot arm. Right: The digital twin which constructed by our method based on the captured data.}
   \label{fig:realrobot}
  \end{figure*}

\paragraph{Environment Configuration}  The model of the RGB-D sensor is Percipio\textregistered FM811-GIX-E1 whose depth capture resolution is 1280 $\times$ 960. The robot arm is STEP\textregistered SR20E which can pick and place box-like \zhn{items} (up to 10kg) with an air pump driven suction cup. Our packing control system is implemented on a desktop computer (\texttt{ubuntu 16.04}), which equips with an \texttt{Intel} \texttt{Xeon} \texttt{Gold} \texttt{5115} CPU @ 2.40 GHz, 64G memory, and \zhn{an} \texttt{Nvidia} \texttt{Titan} \texttt{V} GPU with 12G memory. The size of box items we adopted in our test \zhn{meets} the transportation standard of S.F.Express\textregistered and Taobao\textregistered, which ranges from 20cm to 50cm on each dimension. The items would come continuously and randomly with the conveyor belt at a fixed speed and will be packed by the robot arm to the bin placed on the floor which \zhn{is} driven by our system.

\paragraph{Terminals} Our system contains three terminals. \zhn{The robot} terminal \zhn{drives} the robot arm with Codesys, which is a development environment for programming controller applications. \zhn{The planning} terminal is Python-based, which runs the DRL network we proposed to find the optimal packing strategy. \zhn{The detection} terminal is working with the RGB-D sensor, which \zhn{segments} the input depth image and \zhn{recognizes} the size and location of the coming items. These three terminals are communicated through a TCP protocol which centered on the detection terminal.

\paragraph{Executive Procedure} Before running the whole system, an alignment between our three terminals is required. A hand-eye calibration\cite{dekel2020optimal} is performed between the detection terminal and robot terminal, thus the coordinate system of the object recognized by the RGB-D sensor can be aligned with \zhn{the} robot coordinate system. Meanwhile, we measure the relative position between the pallet  and the robot and align the pallet coordinate system with \zhn{the} robot coordinate system as well.

The detection terminal would capture the depth image of items on the conveyor belt, and \zhn{the} PEAC method\cite{feng2014fast} is performed to judge whether there is an item in the field and measure \zhn{its size}. If an item is detected, the detection terminal would stop the conveyor belt and send the item size $\mathbf{d}_n = [l_n, w_n, h_n]^\top\in\mathbb{Z}^3$ to the planning terminal for packing strategy search.

While the planning terminal is calculating the optimal LP for the input item $n$, the robot terminal would drive the robot arm to pick this item up. It would take few seconds for the robot to complete the picking and the planning terminal can finish the calculating during this period. Once the robot arm leaves the \zhn{camera's} field of view, the detection terminal will restart the conveyor belt and the robot terminal would pack the picked item to the LP given by our planning terminal. At normal speed, it takes \zhn{8-9} seconds for our system to pack a box item from the conveyor belt to the pallet while human palletizers need to spend over 11 seconds without counting the rest time.

\section{Experiments}\label{sec:experiment}
\zhn{Our DRL agent is trained in \texttt{PyTorch}~\cite{paszke2019pytorch}. Training on a spatial resolution of $100 \times 100$ takes about 12 hours and a single decision time is less than $10$~ms.} We compare our method with some state-of-the-art online bin packing methods to demonstrate the superiority. OnlineBPH\cite{ha2017online} and OnlineDRL\cite{zhao2020online} are two representatives, where OnlineBPH is non-learning based and OnlineDRL learns how to pack from the data. \zhn{\cite{zhao2020online} also proposes a heuristic baseline  called \emph{boundary rule} method.  } It replicates human's behavior by trying to place a new item side-by-side with the existing packed items and keep the packing volume as regular as possible.

\zh{
\subsection{Training and Test Set}
}

\zhn{We set $L=W=H=100$ in our experiments with 125 pre-defined item dimensions ($|\mathcal{I}|=125$).  The $100 \times 100 $ resolution is large enough for most of the packing applications in the real world. To avoid over-simplified scenarios, we limit $l_i \leq L/2$, $w_i \leq W/2$, and $h_i \leq H/2$. The training and test sequence is synthesized by generating items out of $\mathcal{I}$, and the total volume of items should be equal to or bigger than \zhn{the} bin's volume. We employ three types of data proposed by \cite{zhao2020online} to train and evaluate our method. One of them is called \textbf{RS} where the sequences are generated by sampling items out of $\mathcal{I}$ randomly. Due to the fact that the optimality of an RS sequence is unknown, the other two types of sequence are generated via \emph{cutting stock}~\cite{gilmore1961linear}. Specifically, items in a sequence are created by sequentially ``cutting'' the bin into items of the pre-defined 125 types so that we understand the sequence may be perfectly packed and restored  to the bin. \textbf{CUT-1} sorts the cut items into the sequence based on $Z$ coordinates of their FLBs, from bottom to top. \textbf{CUT-2} sorts the cut items on their stacking dependency. We generate 2000 sequences on RS, CUT-1, and CUT-2 respectively for testing purposes. The performance of the packing algorithm is quantitated with space utilization (\emph{space uti.}) and the total number of items packed in the bin (\emph{\# items}).
}

\subsection{Ablation Study and Evaluation}

\zhn{Table~\ref{tab:ablation} reports an ablation study about different adoptions. 
The feasibility mask $\mathbf{M}_n$ saves the efforts of exploring invalid actions during the training and guarantees the basic performance of the algorithm. The performance is impaired if the infeasibility loss $E_{\text{inf}}$ is not contributed \zhn{to} the final loss since some exploration would be wasted for infeasible actions. $E_{\text{entropy}}$ encourages the agent to find better solutions and benefits the final performance. The decision condition input (Figure~\ref{fig:3D-BPP} (right)) for each actor-head also facilitates more stable training and better performance.
Figure~\ref{fig:visual_ablation} visualizes the effect of different algorithm settings.
}

\begin{table}[h]
	\centering
	\scalebox{0.9}{
			\begin{tabular}{cccc|c|c}
				\whline{1.15pt}
				{$\mathbf{M}_n$} & {$E_{\text{inf}}$} & {$E_{\text{entropy}}$} & $Cond.$ & {Space uti.} & {\# items} \\
				\whline{0.65pt}
				\xmark & \xmark & \xmark & \xmark & $39.5\%$   & $15.2$  \\
				\xmark & \cmark & \cmark & \cmark & $60.0\%$   & $23.2$  \\
				\cmark & \xmark & \cmark & \cmark & $67.9\%$   & $26.1$  \\
				\cmark & \cmark & \xmark & \cmark & $68.3\%$   & $26.4$  \\
				\cmark & \cmark & \cmark & \xmark & $68.4\%$   & $26.5$  \\
				\cmark & \cmark & \cmark & \cmark & $\bf{71.3\%}$ & $\bf{27.6}$          \\
				\whline{1.15pt}
			\end{tabular}}
\caption{
\zhn{Effect of different adoptions on the RS dataset. $Cond.$ means the decision condition input for each actor-head.}
}\label{tab:ablation}
\end{table}

\begin{figure}[h]
	\centering
	\includegraphics[width = 0.99\linewidth]{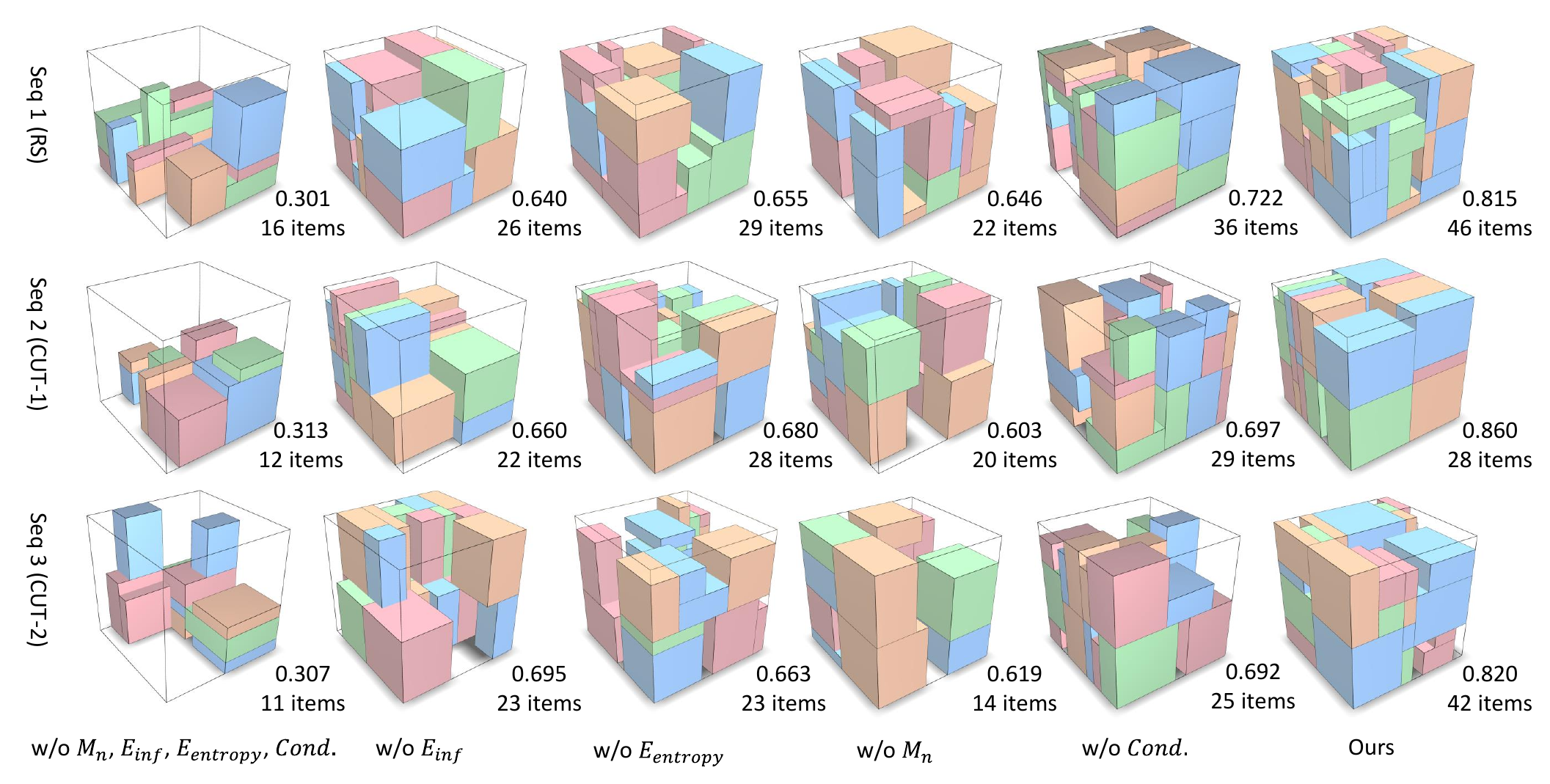} 
	\caption{\zhn{Visualization of the ablation study. The numbers beside each bin are \emph{space uti.} and \emph{\# items}.}	
	}\label{fig:visual_ablation}
\end{figure}

\subsection{Stacking Tree based Stability Estimation}

Next, we demonstrate that the proposed stacking tree based stability estimation is necessary for both efficiency and performance. Table~\ref{tab:stack} reports a quantitated comparison between the proposed method and four alternatives on \zhn{the} RS benchmark. \zhn{The simulator} reports the stability estimation given by Bullet\cite{coumans2015bullet} according to a numerical force analysis. Bullet is able to present precise stability estimation while it is time-consuming. Secondly, \cite{zhao2020online} introduced a \zhn{heuristic-based} approach to discriminate stability of $n$ depends on whether it satisfies any of \zhn{the} following conditions: \zh{1) over $60\%$ of $n$'s bottom area and all of its four bottom corners are supported by existing items; or 2) over $80\%$ of $n$'s bottom area and three out of four bottom corners are supported; or 3) over $95\%$ of $n$'s bottom area is supported.} This approach can also promise $100\%$ accuracy for stability estimation. However, the heuristic constraints \zhn{are} too strict for our network, the packing performance would be limited in this scenario. 

To demonstrate that our stacking tree structure is necessary \zhn{for} stability estimation, we also evaluate the accuracy and performance if only the current packing item $n$'s stability is checked based on our supported centroid approach. It will fail in some cases, which can only achieve $93.7\%$ accuracy. The whole implementation of our stability estimation gives the best packing performance with a $99.9\%$ estimation accuracy  while OnlineBPH\cite{ha2017online} can only achieve $29.1\%$ space utilization with $11.7\%$ estimation accuracy. 
\zhn{Note that our high-speed stability estimation method is for the convenience of DRL training. In practice, we can perform a slower but more accurate stability estimation thread in parallel to avoid misestimation hazards which account for only  $0.1\%$.}

\begin{table}
	\begin{center}
		\scalebox{0.9}{
		\setlength{\tabcolsep}{1.2mm}
		{
		\begin{tabular}{c|c|c|c|c}
			\whline{1.15pt}
			{} & {Space uti.} & {\# items}  & {stability} & {time}      \\
			\whline{0.65pt}
			Bullet simulator\cite{coumans2015bullet}                   & $---$  	 	 & $---$ 		& $---$      & $5.8 \times 10^-2$ s     \\
			Heuristic rule\cite{zhao2020online}              & $58.3\%$  	 & $22.8$ 		& $100.0\%$  & $1.8 \times 10^-4$ s    		\\
			OnlineBPH\cite{ha2017online}                  & $29.1\%$  	 & $11.0$ 		& $11.7\%$   & $---$    	\\
			Only current packing	 			& $66.3\%$     & $25.68$     & $93.7\%$   & $3.8 \times 10^-4 $ s  		\\
			Ours  & $\bf{71.3\%}$  & $\bf{27.6}$  & $99.9\%$   & $5.0 \times 10^-4 $ s			\\
			\whline{1.15pt}
		\end{tabular}}}
	\end{center}
	\caption{Quantitated comparison between the proposed method and four alternatives on RS benchmark. Our method can achieve the best packing performance with the high accuracy of stability estimation. Note that the counting stops when unstable placement occurs.}\label{tab:stack}
	\end{table}


\subsection{Action Space Decomposition}

Benefit from the design of action space decomposition, our model is able to handle the input states with higher resolution. Table~\ref{tab:resolution} reports the packing performance change as the resolution grows. Note that the small fluctuations of performance \zhn{are} normal since different sizes of convolution kernels \zhn{are} adopted to encode features with different \zhn{resolutions}. The packing performance of our method remains stable with different resolutions while the performance of \cite{zhao2020online} drops significantly. The results demonstrate the superiority of the action space decomposition when dealing with high resolution input states.


\begin{table}[h]
\begin{center}
	\scalebox{0.9}{
	\setlength{\tabcolsep}{1.2mm}
	{
	\begin{tabular}{c|c|c|c|c|c|c}
		\whline{1.15pt}
		{} & {} & {$10 \times 10$ } & {$30 \times 30$}  & {$50 \times 50$} & {$100 \times 100$} & {$200 \times 200$}\\
		\whline{0.65pt}
		\multirow{2}*{Ours} & space uti.              & $70.1\%$  	 & $71.7\%$ 	& $72.6\%$  & $71.3\%$  & $70.2\%$   \\
		&\# items                       & $27.1$  	 & $27.7$ 		& $28.1$  & $27.6$  & $27.1$  \\
		\whline{0.65pt}
		\multirow{2}*{\cite{zhao2020online}} & space uti.              & $72.3\%$  	 & $72.4\%$ 	& $51.7\%$  & $--\%$  & $--\%$   \\
		&\# items                       & $27.9$  	 & $27.9$ 		& $20.6$  & $--$  & $--$  \\
		\whline{1.15pt}
	\end{tabular}}}
\end{center}
\caption{Performance comparison with different resolutions between method with decomposed action space (Ours) and method with uniform action space~\cite{zhao2020online}. Note that \cite{zhao2020online} fails with $100 \times 100$ and $200 \times 200$ while our method can still \zhn{maintain} a good performance.}\label{tab:resolution}
\end{table}

We compared our design of action space decomposition with other forms of action space: unified action space and continuous action space. Unified action space refers to the uniform encoding of all actions like $a_n= x_n + L \cdot y_n + L \cdot W \cdot o_n $ and only one actor is trained. Another alternative is to map a unified action space to a continuous domain in the range of [0,1]. As the agent predicts in this continuous domain, we project the result back to the discrete domain. We evaluate these three methods with two different resolutions, $10 \times 10$ and $100 \times 100$. The result is reported in Table~\ref{tab:actionspace}.

\begin{table}
\begin{center}
	\scalebox{0.9}{
	\setlength{\tabcolsep}{1.2mm}
	{
	\begin{tabular}{c|c|c|c|c}
		\whline{1.15pt}
		{} & {} & {Uniform} & {Continuous}  & {Ours}  \\
		\whline{0.65pt}
		\multirow{4}*{$10 \times 10$} & space uti.              				 & $72.3\%$  	 & $54.6\%$ 		& $70.2\%$   \\
		& \# items                                     & $27.9$  	     & $21.5$ 		    & $27.1$     \\
		& GPU Memory                & $948MB$  	 & $942MB$ 		    & $944MB$   \\
		& Network update time & $2.0 \times 10^-2$ s & $1.4 \times 10^-2$ s 	& $2.7 \times 10^-2$ s\\
		\whline{0.65pt}
		\multirow{4}*{$100 \times 100$} & space uti.              				 & $--\%$  	 & $51.2\%$ 		& $71.3\%$   \\
		& \# items                                     & $--$  	     & $20.3$ 		    & $27.6$     \\
		& GPU memory & $8302MB$  	 & $1352MB$ 		& $1356MB$   \\
		& Network update time & $84.7$ s 	& $1.6 \times 10^-2$ s & $5.1 \times 10^-2$ s  \\
		\whline{1.15pt}
	\end{tabular}}}
\end{center}
\caption{Comparison between different \zhn{methods}. The action decomposition method has reached a good level both in performance and training overhead. Unified action space would fail while the resolution increasing to $100 \times 100$.}\label{tab:actionspace}
\end{table}

The unified action space can achieve good performance with high computational efficiency when the state \zhn{resolution} is low. However, the performance drops significantly and the computing overhead has increased dramatically while the action space resolution is high. The continuous form of action space is computational friendly, but cannot achieve a good packing performance. Meanwhile, our action space decomposition method can not only maintain similar performance in different spatial dimensions but also maintain a reasonable computational overhead.

\zhn{

High-resolution action space provides more flexibility to practical packing.
It is very common that the drift exists between the excepted position and actual placement due to robot manipulation error or item misdetection. Here we demonstrate that the high-resolution action space can tolerate item drift more via a toy example in Figure~\ref{fig:resolution}(a)(b).
We also provide quantitative experimental results here.  Assuming that the size of the pallet is 100cm $\times$ 100cm, we will impose a random non-zero drift from -5cm to 5cm to the placed item with a probability of $20\%$. If we pack items at a high resolution of 100 $\times$ 100, 1cm per cell, the packing utilization is $68.2\%$ while the utilization is $71.3\%$ if no drift is imposed. If we execute the same task at a low resolution of 10 $\times$ 10, which means 10cm per cell, the packing utilization drops to $63.1\%$ from $70.1\%$.

If an item's dimension is not an integer with the resolution of the action space, an intuitive approach is filling in this non-unit dimension upwards, e.g., from 15cm to 20cm, which also leads to space waste shown in Figure~\ref{fig:resolution}(c). If the resolution cell is small, the agent can place items in a more compact way, shown in Figure~\ref{fig:resolution}(d). When the item set $\mathcal{I}$ contains non-unit dimension items, our method's performance at 100 $\times$ 100 is $70.9\%$, while $67.0\%$ at 10 $\times$ 10.

}

\begin{figure*}[h]
    \centering
    \includegraphics[width=\linewidth]{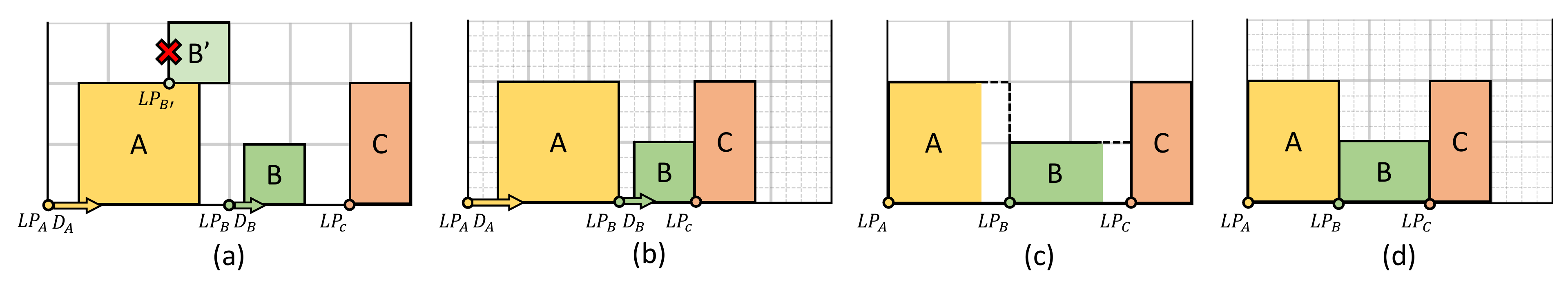}
    \caption{$A$, $B$, and $C$ arrive in order. The grey grids represent the resolution of the action space, items can only be placed on the corners of the grids. (a): Packing in low-resolution action space,  $B$ cannot be placed at $LP_{B'}$ due to drift $D_A$ and placing $B$ at $LP_{B}$ introduces additional space occupancy. The drift $D_B$ will affect the placement of C as well.
	(b): High-resolution packing tolerates item drift and desires less space. (c): Filling up the non-unit dimension items $A$ and $B$ in low-resolution action space takes up five units while (d) packing at a high resolution takes less.}
   \label{fig:resolution}
  \end{figure*}


\zh{
\subsection{Scalability of BPP-$k$}
}
\zhn{Once the agent has mastered the capability of lookahead, it should better exploit the remaining space and deliver a more compact packing. The environment space increases exponentially as $k$ value increases due to the factorial complexity, the agent should make its decision in a reasonable period.}
In Figure~\ref{fig:bppk}(a), we compared the time overhead and algorithm performance of the parallel MCTS compared to the serial approach. In Figure~\ref{fig:bppk}(b), we show that the performance of MCTS improves with the number of lookahead. Note that the root parallelization of MCTS not only improves the efficiency but also \zhn{enables} a larger search space, while the serial MCTS would be easily stuck in a local \zhn{optimum} with a large lookahead size.

\begin{figure}[h]
	\centering
	\begin{overpic}[width = 0.75\linewidth]{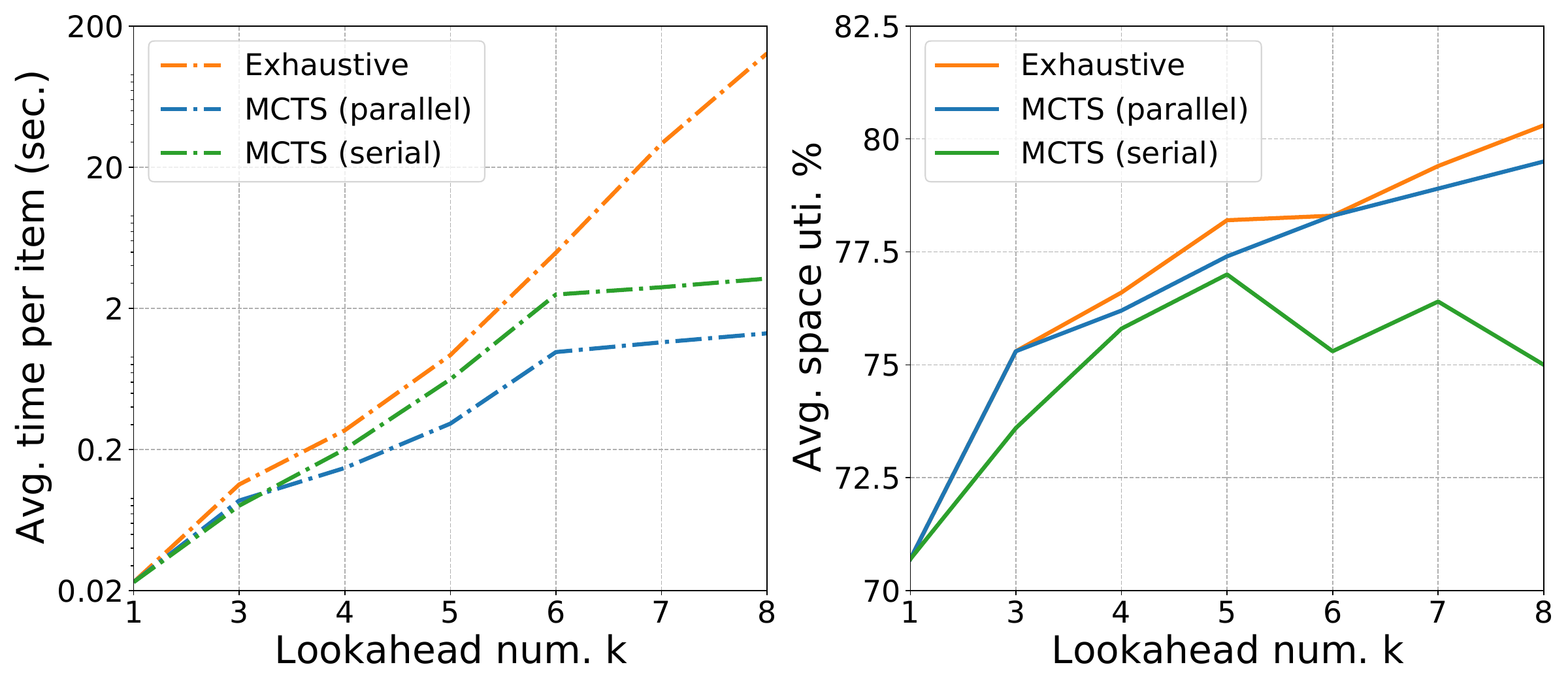}
		\put(25,-2){(a)}
		\put(75,-2){(b)}
	\end{overpic}
	\vspace{2pt}
	\caption{
	\zhn{(a): Our MCTS implementation avoids the factorial computational cost of exhaustive permutation. The parallelization of MCTS also reduces the execution time by nearly half.
	(b) Our parallel MCTS achieves similar performance (avg. space utility) as the brute-force search over permutation tree. The performance of the serial MCTS drops when the lookahead size is larger than 5 due to the ever-increasing search space.}
	}
\label{fig:bppk}
\end{figure}

In the process of MCTS simulation, only \zhn{legally} sampled sequences would be adopted in the training. Sampling quality would directly impact search efficiency. We evaluate 6 different node sampling strategies with $k=10$ and the result is reported in Table~\ref{tab:node}. Here we can see, if we first sample the nodes that actually arrive late with strategy $5^x$, the performance drops significantly since too many illegal sampled sequences \zhn{are} generated in this scenario. If we first sample the nodes that actually arrive early, MCTS begin to be able to utilize lookahead information and achieve better performance. $\mathcal{N}(0,1)$ is the best sampling strategy among them. Fixed sampling order performs worse since it can not explore the search space well.

\begin{table}[h]
\begin{center}
	\scalebox{0.9}{
	\setlength{\tabcolsep}{1.2mm}
	{
	\begin{tabular}{c|c|c|c|c|c|c}
		\whline{1.15pt}
		{} & {$5^x$ } & {$random$}  & {$1/x$} & {$\mathcal{N}(0,1)$} & {$(1/5)^x$}  & {Fixed sampling order} \\
		\whline{0.65pt}
		space uti.              & $72.9\%$  	 & $79.1\%$ 	& $80.6\%$  & $82.5\%$  & $81.2\%$   & $77.9\%$ \\
		\# items                       & $27.7$  	 & $31.1$ 		& $31.6$  & $32.3$  & $31.6$   & $30.0$ \\
		\whline{1.15pt}
	\end{tabular}}}
\end{center}
\caption{The node sample strategy will influence the performance of MCTS, $\mathcal{N}(0,1)$ is the best sampling strategy among them.}\label{tab:node}
\end{table}

\subsection{Comparison with Existing Methods}

We compare to the state-of-the-art non-learning online bin packing method OnlineBPH\cite{ha2017online} to demonstrate that our method can achieve competitive performance on this problem. Note that \cite{ha2017online} allows the agent to select arbitrary one from $k$ lookahead items (i.e., BPP-$k$ with re-ordering) and therefore relaxes the order constraints. We also help this method to make a pre-judgment of stability to make it perform better. In Figure~\ref{fig:bppkcomparison}(a-c), we report the comparison under the setting of BPP-$k$ with $100 \times 100$ state resolution on the three benchmarks. Our method can surpass OnlineBPH\cite{ha2017online} in most cases even re-ordering is allowed in this method while our method does not. Note that OnlineDRL\cite{zhao2020online} which \zhn{adopts} the unified action space would fail in the training with such a high state resolution.

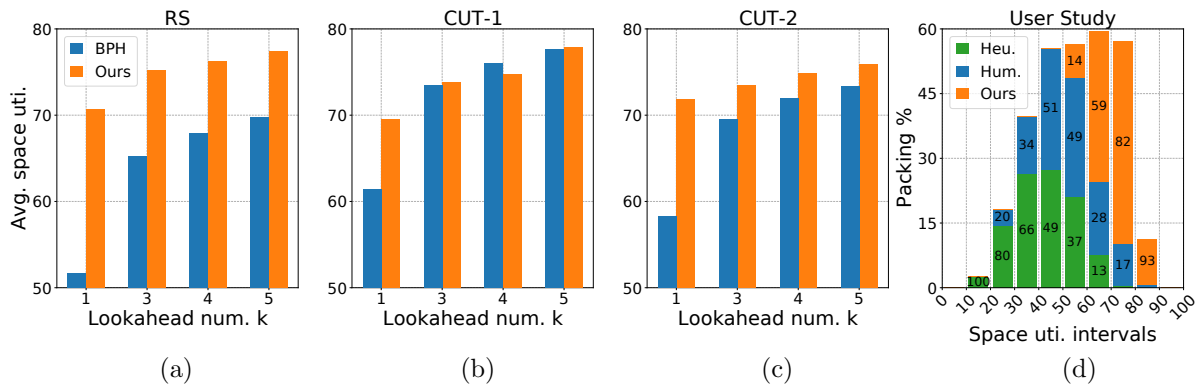
\begin{figure}[h]
	\centering
	\begin{overpic}[width = \linewidth]{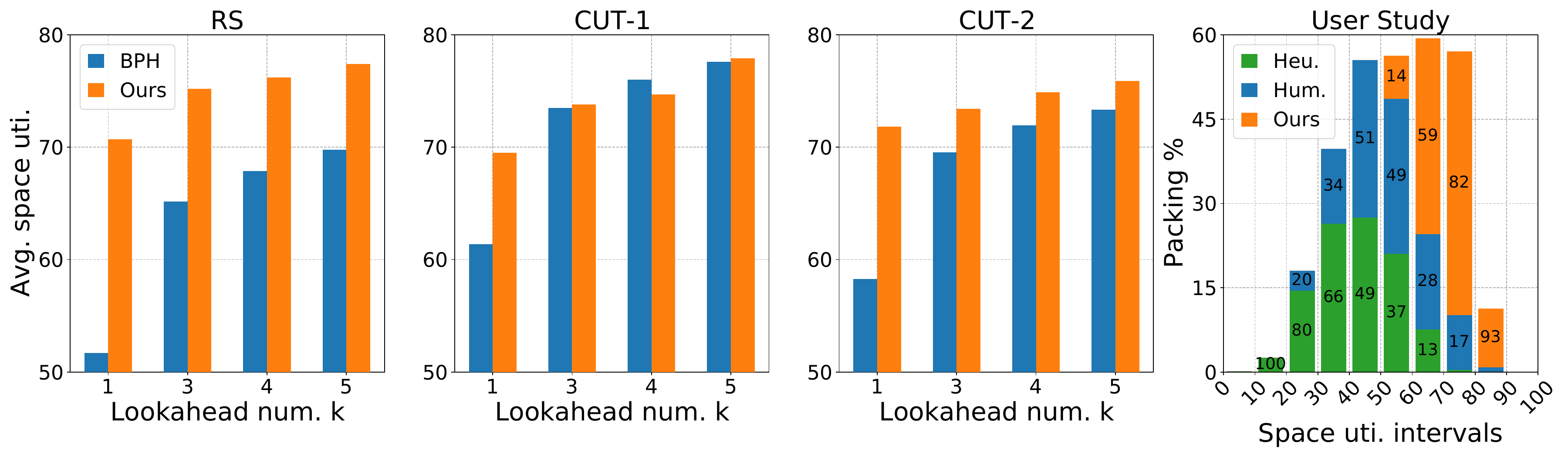}
		\put(13,-2){(a)}
		\put(38,-2){(b)}
		\put(63,-2){(c)}
		\put(88,-2){(d)}
	\end{overpic}
	\vspace{1pt}
	\caption{\kx{(a)-(c): Comparison with the online BPH method~\cite{ha2017online} on BPP-$k$ with $100 \times 100$ state resolution. 
	\zhn{Note that BPH allows selecting any one of the lookahead $k$ items while ours must place them in the order of arrival.
	(d): The distribution of space utilization using boundary rule (Heu.), human intelligence (Hum.), and our BPP-$1$ method (Ours). The more to the right of the distribution, the better the effect of the algorithm.}
	}}\label{fig:bppkcomparison}
\end{figure}

\zhn{The most concerning issue should be the comparison between our algorithm and human intuition. To get the conclusion, we asked 50 human users to pack items manually with a Sokoban-like app proposed by \cite{zhao2020online}. The same sequence is collected and used to test AI (our method). The one with higher space utilization wins. $15$ of the users are palletizing workers and the rest are CS-majored undergraduate/graduate students. No time limit is given. We conduct $2,104$ comparisons and the statistics are plotted in Figure~\ref{fig:bppkcomparison} (d). Our method outperforms human players in general ($1,772$ AI wins vs. $289$ human wins and $43$ evens): it achieves $70.4\%$ average space utilization while human players only have $56.3\%$ (palletizing workers achieve $57.6\%$  while  CS-majored students achieve $55.2\%$). 
}

\begin{figure*}[t!]
	\centering
	\includegraphics[width=0.9\linewidth]{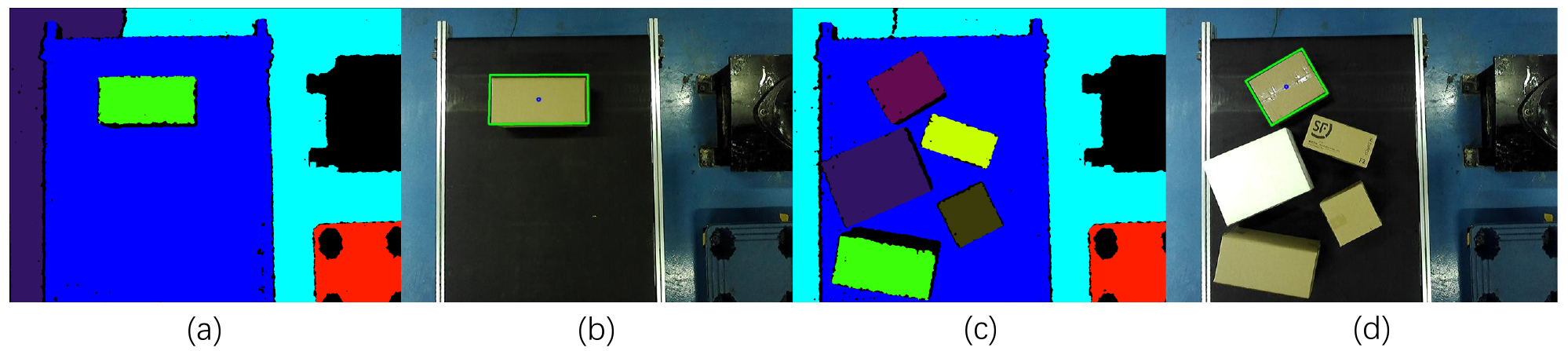}
	\caption{(a): Segmentation result under BPP-$1$ view. (b): The item $n$ is labeled with \zhn{an} anchor, its dimension and location \zhn{are} recognized. (c):  Segmentation result under BPP-$k$ ($k$ is a variable) view. (d) If multiple items are observed by the camera, we sort their positions and select the first one in the queue.}
   \label{fig:view}
  \end{figure*}

\subsection{Robot Implementation in Industry Environment}

We tested our BPP-$1$ and BPP-$k$ method in a practical industry environment, the implementation details can be found in Section~\ref{sec:implementation}. For BPP-$1$, the RGB-D camera \zhn{recognizes} the coming item by segment the captured \zhn{depth map} (Figure ~\ref{fig:view}). If multiple items are observed by the camera at the same time, we sort their positions and select the item at the front of the queue to pack. 

\begin{figure}[h!]
	\centering
	\includegraphics[width = 0.9\linewidth]{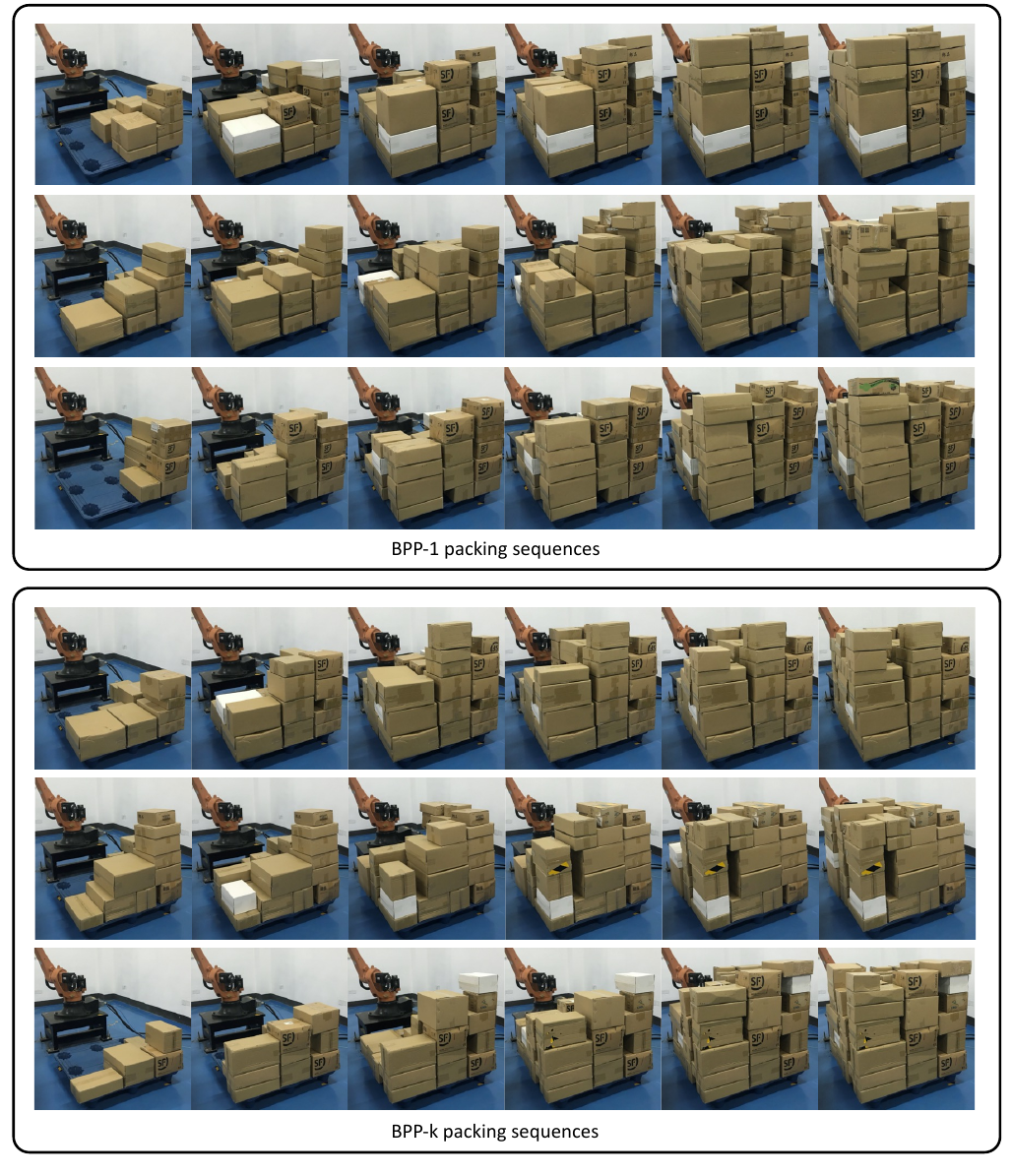} 
	\caption{Visual examples \zhn{are} given by our robot implementation. \zhn{The robot places items in a far-to-near order and reduces collisions with packed items.}}\label{fig:robotvisual}
\end{figure}

Next, we evaluate our BPP-$1$ and BPP-$k$ in this real demo as well in Table~\ref{tab:robot}. Figure~\ref{fig:robotvisual} shows some visual captures of the packing process given by our robot implementation. Note that, as shown in Figure~\ref{fig:view} (d), the number of observed items is uncertain, $k$ is changing while the packing. This is even more challenging than $k$ is fixed. To fully test our BPP-$k$ method, we choose Express\textregistered box standard as our test data since its box size is relatively small which ranges from 20cm to 50cm on each dimension. The input would include more numbers of observed items in this configuration. We also test the packing stability and robot arm collision in this experiment. The packing given by our method is 100$\%$ stable in all the 50 test sequences with BPP-$1$ and BPP-$k$. With the help of our design of collision-free reward, \zhn{the robot places items in a far-to-near order and} only 1 sequence encountered a collision between the robot arm and the packed item with BPP-$1$. If the collision-free reward is not \zhn{included}, \zhn{the agent would place items uniformly on the $XY$ plane of the bin and} the number of collision sequences would significantly increase.


\begin{table}[h]
\begin{center}
	\scalebox{0.9}{
	\setlength{\tabcolsep}{1.2mm}
	{
		\begin{tabular}{c|c|c|c|c|c}
			\whline{1.15pt}
			\multirow{2}{*}{} & \multirow{2}{*}{space uti.} & \multirow{2}{*}{\#items} & \multirow{2}{*}{stability} & \multicolumn{2}{c}{collision rate} \\ \cline{5-6} 
							  &                             &                          &                            & ours   & w/o collision-free reward  \\ \whline{1.15pt}
			BPP-1             & 71.2\%                      & 49.0                     & 100\%                      & 2\%   & 35\%                       \\
			BPP-k             & 77.9\%                      & 53.5                     & 100\%                      & 0\%  & 30\%  \\\whline{1.15pt}                    
			\end{tabular}
	}}
\end{center}
\caption{Performance evaluation of BPP-$1$ and BPP-$k$ in \zhn{a} practical industry environment. Our method can achieve good performance while keeping the stack stable and avoiding the collision.} \label{tab:robot}
\end{table}

\section{Conclusion}
\label{sec:conclusion}

\zhn{We have provided a practical solution to online 3D-BPP with only partial sequence observation and deployed it on a real robot.
}
To learn a feasible packing policy, we propose three new designs. First, we propose an online analysis of packing stability with a novel stacking tree. Second, we propose a decoupled packing policy learning for different dimensions of placement which enables high-resolution discretization and thus high packing precision.
Third, we introduce a reward function \zhn{that} dictates the robot to place items in a far-to-near order and therefore simplifies the collision detection and movement planning of \zhn{the} robotic arm.
As future work, we would like to investigate more on the problem of domain transfer of learned packing policies. In addition, we would like to investigate more relaxations of the problem such as \zhn{stability estimation of items with irregular shapes or uneven mass distribution} and \zhn{ studying} combined solutions of palletizing and depalletizing.



\bibliographystyle{plain}
\bibliography{reference}


\end{document}